# Morphology-Specific Closed-Loop Control of Logarithmic-Spiral Continuum Arms via Online Jacobian Error Compensation

Partha Datta[1,2,†], Yi Jin[1,†], Wei Lin[2], C. Chase Cao[1,2,3,*]

[1]*Laboratory for Soft Machines and Electronics, Department of Mechanical and Aerospace Engineering, Case Western Reserve University, Cleveland, OH 44106, USA*

[2]*Department of Electrical, Computer and Systems Engineering, Case Western Reserve University, Cleveland, OH 44106, USA*

[3]*Advanced Platform Technology (APT) Center, Louis Stokes Cleveland VA Medical Center, Cleveland, OH 44106, USA*

[†]Equal contribution. [*]Corresponding author: C. Chase Cao (ccao@case.edu)

**Abstract**

Logarithmic spirals are ubiquitous in biological appendages and provide an attractive morphology for continuum manipulators capable of reaching, wrapping, and grasping. Recently reported logarithmic-spiral robots demonstrated scalable fabrication and versatile grasping but lacked inverse kinematics and closed-loop control. This work presents the first morphology-specific closed-loop task-space control framework for logarithmic-spiral continuum arms. A segmented tendon-driven model with a centerline backbone and equilateral tendon routing is developed in MuJoCo to capture tapered compliance and contact dynamics. An analytical task-space Jacobian is derived directly from the logarithmic-spiral kinematics and combined with online Jacobian error compensation using a Broyden secant update and Kalman-filter estimation. The resulting controller continuously corrects modeling errors arising from nonlinear deformation, contact, and geometric mismatch. The framework is validated through planar and spatial simulations, including trajectory tracking, attitude regulation, disturbance rejection, three-dimensional position tracking, and simultaneous position-orientation control. Compared with a piecewise-constant-curvature (PCC) baseline, the proposed method consistently reduces tracking errors, suppresses attitude drift, and maintains a bounded Jacobian estimation error. The controller is further applied to morphology-enabled manipulation tasks, including obstacle-assisted reach-wrap-release motions, adaptive whole-arm grasping, and cooperative multi-arm object handling. Results demonstrate that combining logarithmic-spiral morphology with online Jacobian compensation enables accurate, robust, and scalable control of highly underactuated continuum manipulators. The proposed framework establishes a physics-grounded baseline for future hardware implementation and learning-augmented soft robotic control.

## 1. Introduction

Soft continuum manipulators offer a compelling alternative to rigid-link robots for manipulation in cluttered and uncertain environments [1], [2], [3]. A broad range of soft arms, hands, and grippers has been demonstrated, including cable-driven manipulators with variable stiffness for safe human-machine interaction [4], pneumatically actuated humanoid hands with large grasping force [5], self-powered soft actuators and grippers [6], fabric-based omnidirectional bending actuators with tunable stiffness [7], and tensegrity-inspired lightweight robotic arms [8]. Bioinspiration has long driven soft-robot design more broadly, producing worm-inspired crawlers for confined-space inspection [9], [10], anisotropic-hydrogel soft actuators and robots [11], insect-scale untethered propulsors [12], and liquid-metal-integrated soft systems [13], with magnetically actuated small-scale soft robots adding further morphing capability [14]. Across compliant biological appendages, from seahorse tails to elephant trunks and octopus arms (**Fig. 1A**), a recurring geometric motif is the logarithmic spiral: a tightly packed curl whose radius of curvature varies smoothly along the body, uncurling to reach and curling to grasp. Rather than emulating any single organism, recent work has argued that this spiral is a design principle, decoupling morphology from biology and enabling a single discretized geometry to be scaled across more than two orders of magnitude [15].

Continuum-robot modeling spans piecewise-constant-curvature (PCC) approximations [16], variable- and polynomial-curvature models, and physically grounded Cosserat-rod formulations that capture bending, shear, torsion, and extension along the backbone [17], [18]. Recent extensions add cross-sectional deformation [19] and discrete elastic-rod variants that incorporate state observers for underactuated continuum robots [20]. PCC is computationally efficient, but its circular-arc assumption limits accuracy, particularly under contact and friction. Cosserat-rod and finite-element models are accurate but computationally costly [21]. Tendon-driven designs are favored for compact actuation and high achievable curvature [22], [23].

On the control side, three families are relevant [24]. Model-based feedback and inverse-kinematic methods exploit analytical or learned Jacobians. Giorelli et al. [25] derived an inverse-kinematic Jacobian for a cable-driven arm under a PCC approximation, and dynamic model-based controllers have been developed for soft arms [26]. Optimization-based methods formulate task tracking with explicit stability and actuator constraints, but can be computationally demanding due to the many degrees of freedom in a continuum backbone. Online and model-free methods estimate the input-to-task mapping from data, including learning-based [27] and Koopman-operator [28] approaches, as well as Kalman-filter Jacobian-error estimation [21], which improves tracking and disturbance rejection without offline training.

Wang et al. [15] introduced SpiRobs, a class of cable-driven logarithmic-spiral robots that replicate octopus-like curling and uncurling using two or three cables. They demonstrated grasping across a wide range of sizes and weights, and reported miniaturized and meter-scale variants as well as multi-arm arrays. Their design is generated directly from the spiral parameters, enabling fast, repeatable fabrication. However, they relied on manual or piecewise-linear open-loop sequencing and explicitly identified the absence of an inverse-kinematic and dynamical model, and the consequent lack of feedback control, as the principal limitation and top priority for future work. The SpiRobs design is distinguished by controllable spiraling at any point along the body, scalable fabrication, and an octopus-inspired open-loop grasping strategy. Related rigid-soft synergies have reproduced elephant-like spiral grasping [29]. This work aims to address that gap by contributing a closed-loop task-space controller built on the analytical spiral kinematics. To our knowledge, closed-loop task-space control tailored to this logarithmic-spiral morphology, together with closed-loop execution of its grasping sequence, has not been reported previously.

Controlling such arms is difficult because of high-dimensional shape dynamics, nonlinear tendon routing, friction, and strong coupling between actuation and contact [2]. The spiral arm is also severely underactuated, with only two or three cables commanding a backbone of many segments. As a result, the input-to-task mapping is sensitive to model mismatch and contact. Recently, Zhai et al. [21] showed, on a tendon-driven planar continuum manipulator, that a piecewise-constant-curvature (PCC) Jacobian augmented with an online Jacobian error estimate from a Kalman filter substantially improves task-space tracking without offline training; we adopt this estimation principle but use the analytical logarithmic-spiral kinematics, rather than a generic constant-curvature arc, as the nominal model. This motivates the central question of our work: can a lightweight, model-based online-Jacobian controller, tailored to the spiral geometry, achieve reliable closed-loop task-space tracking and grasping on this highly underactuated platform?

In this paper, we develop a morphology-specific task-space control framework for tendon-driven logarithmic-spiral continuum arms (**Fig. 1**). The main contributions are:

**1)** *A segmented, physics-based MuJoCo model of a logarithmic-spiral continuum arm with a centerline backbone and equilateral spatial tendon routing.* The model captures the tapered bending behavior of the logarithmic-spiral arm and is validated against a Cosserat-rod analytical formulation.

**2)** *A task-space controller based on an analytical Jacobian derived directly from logarithmic-spiral kinematics.* Unlike PCC models, the proposed formulation preserves the continuously varying curvature and distal geometric growth of the spiral morphology. The nominal Jacobian is further enhanced through online error compensation using a Broyden secant update and Kalman-filter-based Jacobian error estimation, enabling correction of residual contact and modeling errors relative to the geometry-exact spiral model.

**3)** *Comprehensive validation across trajectory tracking, attitude regulation, disturbance rejection, spatial position-orientation tracking, and morphology-enabled manipulation tasks.* These studies include closed-loop reaching behaviors that transition naturally into reach–wrap–hold grasping, as well as cooperative multi-arm entanglement grasping and object transport. Quantitative comparisons against PCC and open-loop baselines demonstrate improved tracking accuracy, reduced attitude drift, enhanced robustness to disturbances, and more reliable manipulation performance. To the authors' knowledge, this is the first closed-loop control framework developed specifically for logarithmic-spiral continuum manipulators and the first systematic evaluation of their morphology-enabled manipulation capabilities.

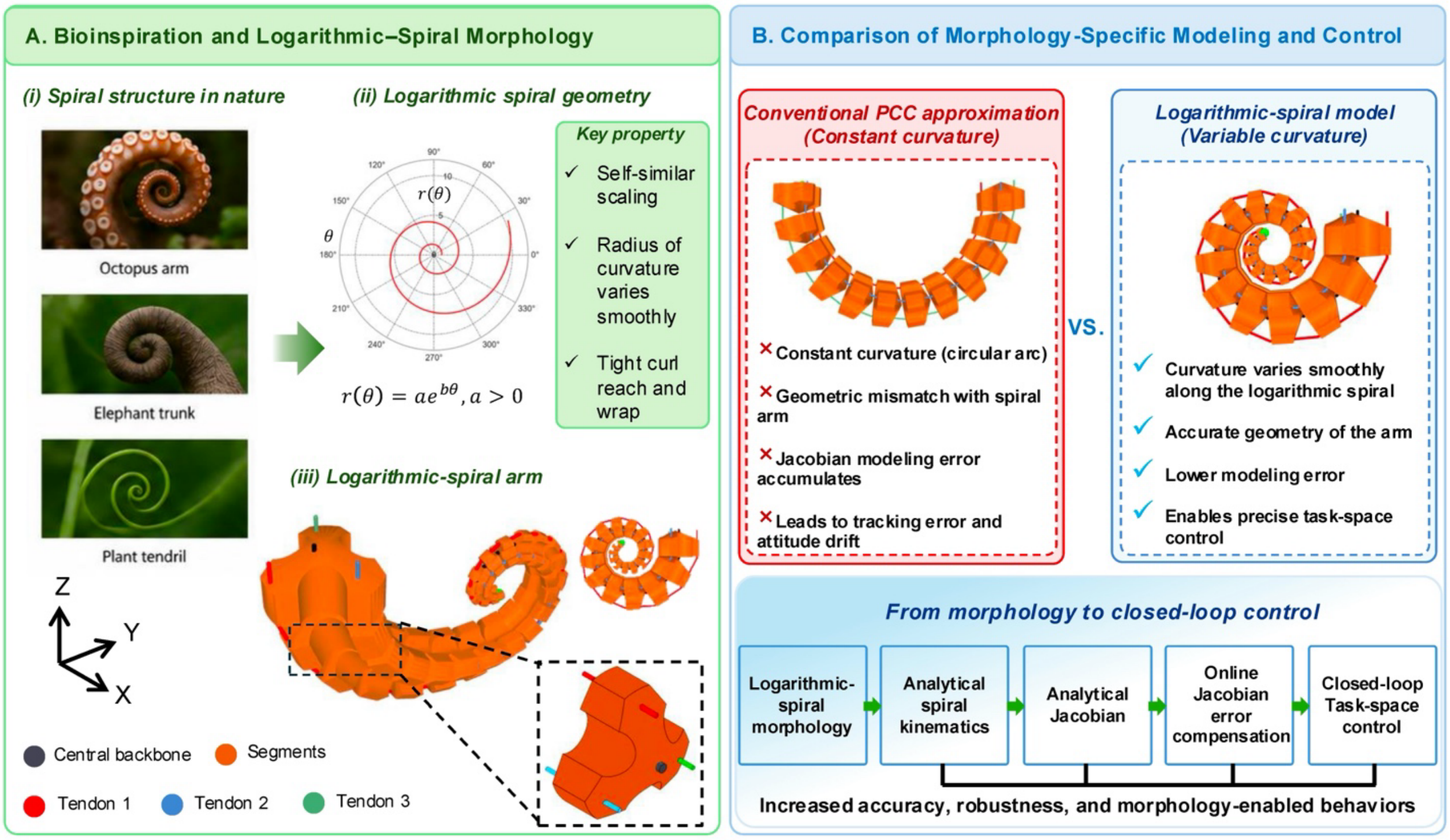

**Figure 1. Concept and motivation of the proposed logarithmic-spiral continuum arm and morphology-specific control framework. (A)** Spiral geometries commonly observed in nature (e.g., octopus arms, elephant trunks, and plant tendrils) inspire the tendon-driven continuum arm, which is generated from a logarithmic spiral $r(\theta) = ae^{b\theta}$ and features a central backbone with equilateral tendon routing. The resulting morphology exhibits smoothly varying curvature and enables reach–wrap–hold behaviors. **(B)** Comparison between conventional PCC modeling and the proposed logarithmic-spiral model. The spiral model captures the variable-curvature morphology more accurately, enabling analytical kinematics, morphology-specific Jacobian modeling, online Jacobian error compensation, and improved closed-loop task-space control.

## 2. Design and Modeling of the Logarithmic-Spiral Arm

We adopt a logarithmic-spiral design for the arm morphology in closed-loop control. The backbone centerline in the plane is a logarithmic spiral

$$r(\theta) = a\, e^{b\theta}, \qquad a > 0 \tag{1}$$

where $a > 0$ is a scaling factor and $b$ is the cotangent of the constant polar tangential angle. Converting to Cartesian coordinates yields the backbone point

$$\mathbf{p}(\theta) = \begin{bmatrix} x(\theta) \\ y(\theta) \end{bmatrix} = \begin{bmatrix} a\, e^{b\theta} \cos\theta \\ a\, e^{b\theta} \sin\theta \end{bmatrix} \tag{2}$$

The discrete units are obtained by discretizing and mirroring the spiral about a central axis, so that lengths are measured along this central axis [9]. To attach an end-effector frame, we align the local x-axis with the spiral tangent. For a logarithmic spiral, the angle between the radius vector and the tangent is constant,

$$\psi = \arctan\left(\frac{1}{b}\right) \tag{3}$$

and the tangent heading angle is

$$\alpha(\theta) = \theta + \psi \tag{4}$$

The corresponding planar rotation matrix is

$$\mathbf{R}(\alpha) = \begin{bmatrix} \cos\alpha & -\sin\alpha \\ \sin\alpha & \cos\alpha \end{bmatrix} \tag{5}$$

the homogeneous transformation (pose) of the backbone-attached frame at spiral parameter $\theta$ is

$$\mathbf{T}(\theta) = \begin{bmatrix} \mathbf{R}\big(\alpha(\theta)\big) & \mathbf{p}(\theta) \\ \mathbf{0}^{\top} & 1 \end{bmatrix} = \begin{bmatrix} \cos(\theta+\psi) & -\sin(\theta+\psi) & a\, e^{b\theta}\cos\theta \\ \sin(\theta+\psi) & \cos(\theta+\psi) & a\, e^{b\theta}\sin\theta \\ 0 & 0 & 1 \end{bmatrix} \tag{6}$$

### 2.1 Per-segment kinematic formulation and degrees of freedom

We model the arm as n logarithmic-spiral segments connected in series. Segment $i$ is a chain of $N_i$ rigid teeth ($j = 0, \dots, N_{i-1}$, from segment base to tip) joined by compliant bending joints, with self-similar taper

$$s_{i,j} = s_{i,0}\gamma_i^{j}, \quad d_{i,j} = d_{i,0}\gamma_i^{j}, \quad k_{i,j} = k_{i,0}\gamma_i^{3j}, \quad \gamma_i \in (0,1) \tag{7}$$

where $s_{i,j}$ is the tooth length, $d_{i,j}$ is the cable moment arm (the perpendicular distance from the joint bending axis to the cable routing path, which sets the torque a cable tension produces), and $k_{i,j}$ is the joint bending stiffness; $k_{i,j} \propto \frac{EI}{s_{i,j}}$ with $I \propto d_{i,j}^4$ and $s_{i,j} \propto d_{i,j}$ gives $k_{i,j} \propto d_{i,j}^3$. The taper ratio and per-joint notch limit $\beta_i$ follow from the spiral growth rate via $\gamma_i = e^{-k_{i,j}\beta_i}$. Each segment carries three independent, unilateral cables at $\psi_{i,m} = \{0, 2\pi/3, 4\pi/3\}$ with tensions $T_{i,m} \geq 0$, producing a resultant moment

$$\widehat{M}_i = \sum_{m=1}^{3} T_{i,m} \left(\cos\psi_{i,m}, \sin\psi_{i,m}\right) \in \mathbb{R}^2, \quad T_{i,m} \geq 0 \tag{8}$$

a rank-2 map from three tensions onto a two-dimensional curvature command (the common mode $T_{i,1} = T_{i,2} = T_{i,3}$ is its kernel). With a linear elastic joint, the bend angles follow the exponential, saturating profile

$$\theta_{i,j} = \min\left(\beta_i,\ A_i {\gamma_i}^{-2j}\right), \quad A_i = \frac{d_{i,0}}{k_{i,0}} \parallel \widehat{M}_i \parallel \tag{9}$$

the curvature grows toward the distal tip — the logarithmic-spiral signature that a constant-curvature (PCC) model cannot represent — and the distal joints saturate first, reaching the maximal curl $\Phi_i = N_i \beta_i$ at $A_i = \beta_i$. We encode each segment by its curvature vector

$$u_i = (A_i \cos\varphi_i,\ A_i \sin\varphi_i) \in \mathbb{R}^2 \tag{10}$$

which is smooth through the straight pose and removes the bending-plane chart singularity of a raw $(\theta, \varphi)$ parametrization. Composing the per-segment transforms $T_i(u_i)$ gives the forward kinematics and configuration

$$T(q) = \prod_{i=1}^{n} T_i\,(u_i), \quad q = (u_1, \dots, u_n) \in \mathbb{R}^{2n} \tag{11}$$

so $\dim Q = 2n$. Partitioning the tip transform T into its rotation block R and translation block p, the tip position is $p$ and the tip tangent is $t$, the third column of R (the tip-frame z-axis). The task vector is $x = (x, y, z, \theta)$, with elevation $\theta = atan2\left(\sqrt{\left({t_x}^2 + {t_y}^2\right)}, t_z\right)$. This is robust even when the azimuth degenerates near a vertical tangent. Because a tendon chain cannot generate roll, the controllable pose coordinates are five $\{x, y, z, \theta, \alpha\}$, and the number of controllable task degrees of freedom is

$$d(n) = \min(2n, 5) \tag{12}$$

Thus, a single segment gives $d = 2$; the two-segment arm used here gives $d = 4$ ($x, y, z, \theta$; tip azimuth and roll left free); and three segments would give the full $d = 5$. The nominal task Jacobian is therefore $\mathbf{J}_q = \partial f / \partial q$, obtained by differentiating this forward map; the full statics, compliance, and conditioning derivation is given in the Supplementary Material (Section S2). The table below contrasts the single- and two-segment cases.

**Table 1.** Comparison of reachable sets and singularity structure for one versus two segments (single- vs. two-segment structure of the logarithmic-spiral arm).

| | **Single segment (n = 1)** | **Two segments (n = 2)** |
|---|---|---|

| Configuration | $u_1 \in \mathbb{R}^2$ (2 DOF) | $(u_1, u_2) \in \mathbb{R}^4$ (4 DOF) |
|---|---|---|
| Task DOF, d = min(2n,5) | 2 | 4 |
| Controllable coords | 2 of {x, y, z, θ, α} | x, y, z, θ |
| Shape | planar (one plane) | generally spatial (two planes) |
| Reachable set | 2-D surface of revolution | 4-D submanifold of SE(3) |
| Tip orientation | slaved to position | elevation independent of position |
| Singular loci | home; saturation | + coplanar $\varphi_1 \equiv \varphi_2$ (mod π) |

### 2.2 Taper geometry and backbone length

Sections 2.2 and 2.3 specialize the general per-segment formulation of Section 2.1 to the explicit single-spiral parameterization used by the controller. Specifically, the taper-angle and backbone-length relations, together with the closed-form inverse in Section 2.3, furnish the tendon-length-to-configuration map. Composing this map with the forward kinematics of Eq. (6) yields the nominal task Jacobian $J_q$ used in Section 3. The taper angle of the discretized spiral units is related to the spiral parameters by

$$\phi = 2\arctan\left[\frac{b\left(e^{2\pi b}-1\right)}{\sqrt{b^2+1}\left(e^{2\pi b}+1\right)}\right] \tag{13}$$

which is independent of θ, confirming that the expanded body is uniformly tapered. The backbone arc length along the central spiral from the virtual tip $(\theta = 0)$ to a point at parameter $\theta_0$, corresponding to the commanded tendon displacement, is given by

$$L(0,\theta_0) = \frac{a\sqrt{b^2+1}}{b}\left(e^{b\theta_0}-1\right) \tag{14}$$

### 2.3 Inverse length-to-configuration mapping

To recover the configuration from a commanded backbone length, we use the differential arc-length relation

$$\frac{dr}{d\theta} = b\,r(\theta), \qquad ds = \sqrt{r(\theta)^2 + \left(\frac{dr}{d\theta}\right)^2}\; d\theta \tag{15}$$

Solving for the spiral parameter θ, which encodes the tip configuration, gives the inverse mapping

$$\theta = \frac{1}{b}\ln\left(e^{b\theta_0} + \frac{bL}{a\sqrt{b^2+1}}\right) \tag{16}$$

This closed-form relation seeds the analytical Jacobian used by the controller in Section 3.

### 2.4 Segmented MuJoCo model and tendon routing

The continuous spiral is discretized with a fixed angular step into tapered rigid units connected by compliant elastic joints. The planar validation arm uses two antagonistic tendons, while the spatial arm uses three tendons in an equilateral routing pattern that enables bending in any plane. A MuJoCo [30] model replicates the segmented structure, the equilateral tendon routing, the elastic restoring forces, and the frictional contact. Because the geometry, joint compliance, and routing are generated procedurally from the spiral parameters, the model provides a repeatable simulation baseline for controller development and grasping evaluation.

### 2.5 Compliance and bending-stiffness profile

As established in Section 2.1, the spiral taper produces graded compliance: proximal segments are stiff, and distal segments are highly compliant, providing the mechanical basis for distal curling and proximal load support. Assigning each inter-segment joint a rotational stiffness and computing an equivalent bending stiffness yields a profile that decreases monotonically from base to tip. The bending shape produced by the segmented MuJoCo model matches the analytical spiral predicted by Cosserat-rod theory [22] (Supplementary Fig. S1), confirming that the discretized model preserves the intended continuum mechanics.

## 3. Control Methodology

Let the tendon inputs be $\mathbf{u}$ and the measured task-space pose be $\mathbf{x}$ (planar position and attitude). Over small-time steps, the kinematics are locally linearized as

$$\Delta\mathbf{x}_k \approx \mathbf{J}_k\,\Delta\mathbf{u}_k \tag{17}$$

where $\mathbf{J_k}$ is the instantaneous task-space Jacobian. The nominal $\mathbf{J_k}$ is obtained analytically by differentiating the segmented spiral task map of Section 2, $\mathbf{J} = \partial \mathrm{f}/\partial \mathrm{q}$ with $\mathbf{x} = \mathbf{f(q)}$, composed with the tendon-to-configuration map, rather than from a generic constant-curvature arc.

### 3.1 Incremental control (damped least squares)

The task Jacobian J used here is a displacement (incremental) Jacobian: it maps actuator increments to task-space pose increments rather than joint velocities to tip velocities. Given a desired incremental motion $\mathbf{\Delta x}^{\mathrm{des}} = \mathbf{x}^{\mathrm{ref}} - \mathbf{x}$, the input update is computed by damped least squares,

$$\Delta\mathbf{u}_k = \left(\mathbf{J}_k^{\top}\mathbf{J}_k + \lambda^2\mathbf{I}\right)^{-1}\mathbf{J}_k^{\top}\Delta\mathbf{x}_k^{\mathrm{des}} \tag{18}$$

with damping $\lambda > 0$ for robustness near singularities, and saturated to the actuator limits,

$$\mathbf{u}_{k+1} = \mathrm{clip}(\mathbf{u}_k + \Delta\mathbf{u}_k,\ \mathbf{u}_{\min},\ \mathbf{u}_{\max}) \tag{19}$$

### 3.2 Online Jacobian update (Broyden)

Because the analytical Jacobian is inaccurate under contact and model mismatch, an estimated Jacobian is updated from measured increments,

$$\Delta \mathbf{x}_k = \mathbf{x}_{k+1} - \mathbf{x}_k, \qquad \Delta \mathbf{u}_k = \mathbf{u}_{k+1} - \mathbf{u}_k \tag{20}$$

using the rank-one Broyden secant update

$$\mathbf{J}_{k+1} = \mathbf{J}_k + \frac{(\Delta \mathbf{x}_k - \mathbf{J}_k \Delta \mathbf{u}_k)\Delta \mathbf{u}_k^{\top}}{\Delta \mathbf{u}_k^{\top} \Delta \mathbf{u}_k + \eta} \tag{21}$$

where $\eta > 0$ regularizes the denominator against vanishing input increments.

### 3.3 Kalman-filter Jacobian error compensation

Following Zhai et al. [21], a more stable alternative treats the analytical Jacobian as a nominal model and estimates an additive Jacobian-error term online with a Kalman filter, using real-time actuation and pose measurements:

$$\tilde{\mathbf{J}} = \mathbf{J}_{\text{model}} + \delta\hat{\mathbf{J}}, \qquad \mathbf{r}_k = \Delta \mathbf{x}_k - \mathbf{J}_k \Delta \mathbf{u}_k \tag{22}$$

Under a quasi-static assumption, the Jacobian error evolves linearly with additive noise and is estimated with a linear Kalman filter; constraining successive estimates reduces fluctuations and improves stability near singularities and under contact. Here, the additive term is a modeling assumption on the evolution of the linearization (Broyden) residual, not measurement noise injected into the simulator: the filter regularizes the online Jacobian-error estimate rather than denoising sensor readings, and tip poses are read exactly from the simulator. Throughout the results, the online Jacobian-error metric is $\mathbf{e_J} = \|\tilde{\mathbf{J}} - \mathbf{J}^*\|_{\mathbf{F}}$, the Frobenius norm of the difference between the controller's working Jacobian $\tilde{\mathbf{J}}$ (Eq. (22)) and the true plant Jacobian $\mathbf{J}^*$ evaluated numerically from the simulator; the online compensation drives $\mathbf{e_J}$ toward zero, whereas omitting it lets $\mathbf{e_J}$ grow as the nominal model drifts from the plant. The complete closed-loop architecture, integrating the analytical spiral model, the Broyden update, and the Kalman-filter error compensation, is shown in **Figure 2**.

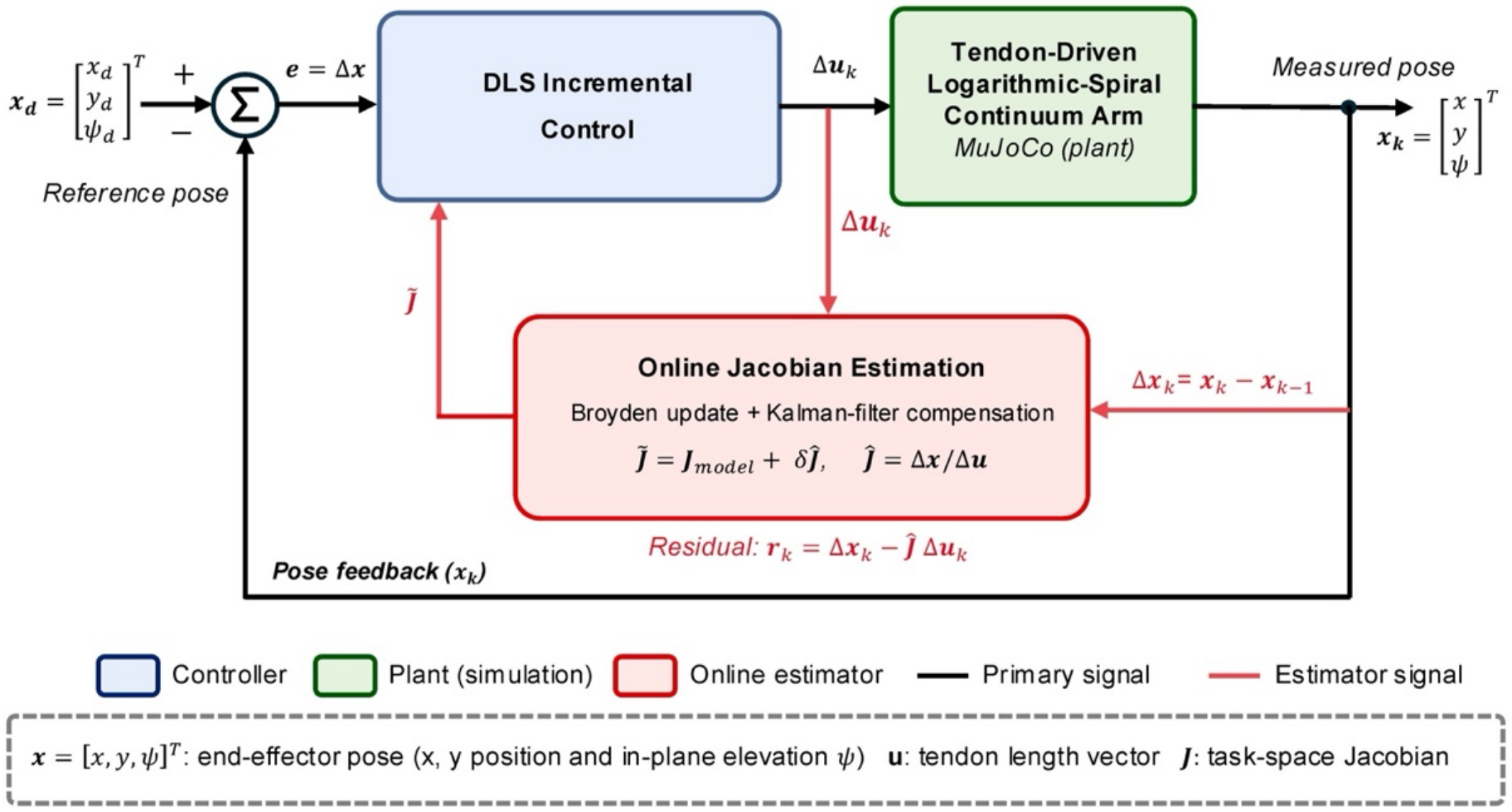

**Figure 2. Closed-loop task-space control architecture for the logarithmic-spiral continuum arm.** The reference pose $\mathbf{x}_d = [x_d, y_d, \psi_d]^T$ is compared with the measured pose $\mathbf{x}_k = [x, y, \psi]^T$ to generate the task-space error $\mathbf{e} = \Delta\mathbf{x}$. A damped least-squares (DLS) incremental controller computes the tendon command increment $\Delta\mathbf{u}_k$, which is applied to the tendon-driven logarithmic-spiral continuum arm simulated in MuJoCo. An online Jacobian estimation module combines a Broyden secant update with Kalman-filter-based error compensation to continuously update the task-space Jacobian, $\tilde{\mathbf{J}} = \mathbf{J}_{model} + \delta\hat{\mathbf{J}}$,using measured pose changes and actuation residuals. The compensated Jacobian is fed back to the controller, enabling accurate and robust closed-loop tracking in the presence of model uncertainty and external disturbances.

### 3.4 Closed-loop algorithm

The task-space tracking procedure with online compensation is summarized in **Algorithm 1**.

**Algorithm 1 Task-space tracking with online Jacobian error compensation**

**Input:** reference poses $\{x_k^{ref}\}$; initial input $u_0$; analytical Jacobian $\boldsymbol{J_{model}}$; damping $\lambda$; regularizer $\eta$

**Output:** tendon commands $\{u_k\}$ tracking the reference trajectory

1: $\hat{\boldsymbol{J}} \leftarrow \boldsymbol{J_{model}};\ \ \delta\hat{\boldsymbol{J}} \leftarrow 0$

2: **while** task not complete **do**

3: $x \leftarrow measurePose();\ \ \Delta x_{des} \leftarrow x_k^{ref} - x$

4: $\tilde{\boldsymbol{J}} \leftarrow \boldsymbol{J_{model}} + \delta\hat{\boldsymbol{J}}$ ▷ *compensated Jacobian*

5: $\Delta u \leftarrow \left(\tilde{\boldsymbol{J}}^T\tilde{\boldsymbol{J}} + \lambda^2\boldsymbol{I}\right)^{-1}\tilde{\boldsymbol{J}}^T\,\Delta x_{des}$ ▷ *damped least squares*

6: $u \leftarrow clip(u + \Delta u, u_{min}, u_{max});\ \ apply\ u;\ \ x' \leftarrow measurePose()$

| | | |
|---|---|---|
| 7: | $\Delta x \leftarrow x' - x; \; r \leftarrow \Delta x - \hat{\boldsymbol{J}}\,\Delta u$ | ▷ *residual* |
| 8: | $\hat{\boldsymbol{J}} \leftarrow \hat{\boldsymbol{J}} + r\,\Delta u^{\top} \;/\; (\Delta u^{\top}\,\Delta u + \eta)$ | ▷ *Broyden secant update* |
| 9: | $\delta\hat{\boldsymbol{J}} \leftarrow KalmanUpdate(\delta J, r, \Delta u)$ | ▷ *Kalman error compensation* |
| 10: | **end while** | |

**4. Grasping Strategy**

The grasping strategy is inspired by octopus arms, which achieve versatile manipulation through coordinated reaching, wrapping, and holding behaviors [31]–[35]. The logarithmic-spiral arm progressively uncurls to reach a target and subsequently re-curls to envelop and secure the object. Accordingly, grasping is divided into three stages: **reaching**, **wrapping**, and **holding**. During reaching, the controller regulates the arm configuration to position and orient the tip toward the target. Once contact is detected through increased actuator effort, the arm transitions to the wrapping stage, where its curvature increases and the compliant spiral body conforms to the object surface to generate a whole-arm enveloping grasp. The holding stage maintains stable object retention through passive compliance and distributed contact forces.

A key feature of this strategy is the division of labor between control and morphology. Unlike conventional grasping approaches that require continuous feedback regulation throughout the manipulation process [36][37], the proposed controller is responsible only for the reaching phase. Wrapping and holding emerge naturally from the logarithmic-spiral geometry, passive compliance, and contact mechanics of the arm. This morphology-assisted behavior allows the manipulator to adapt to objects of different sizes and shapes with minimal control complexity. In obstacle-rich environments, the arm can further exploit environmental support to conform around barriers and achieve robust grasps. Whereas the original SpiRobs system relied on open-loop actuation sequences [9], the present work closes the loop on the reaching phase using the task-space controller described in Section 3. The framework supports both single-arm whole-body wrapping and cooperative multi-arm entanglement grasps [37], providing a foundation for morphology-enabled manipulation in highly underactuated continuum robotic systems.

**5. Simulation Experiments and Results**

We validate the controller in MuJoCo through four experiments, followed by grasping demonstrations and a model-validation study reported in the Supplementary Material. Experiments 1 and 2 use the planar two-tendon arm; Experiment 3 (disturbance rejection) also uses the planar arm; and Experiment 4 uses the spatial three-tendon arm. Each experiment defines a reference, fixes variables, measures outcomes, and establishes baselines. Each experiment compares the proposed method (analytical spiral Jacobian with online Broyden-plus-Kalman compensation) against a PCC baseline: a piecewise-constant-curvature nominal model that substitutes for the spiral Jacobian and runs without online compensation, isolating the contribution of the spiral-specific kinematics. The comparison tables report two reference conditions alongside the proposed controller: a PCC baseline that uses a nominal constant-curvature model without the online compensation, and an open-loop condition that executes a pre-computed tendon command without pose feedback. The disturbance experiment (Experiment 3) reports the response of the proposed controller only and does not include a baseline comparison. The simulation is deterministic—fixed model, initial state, control sequence, and solver and engine settings, with no injected sensor or process noise—so a repeated run reproduces the exact same result. For each task-space quantity, we report two measures, expressed as RMSE/MaxAE: the root-mean-square error and the maximum absolute error over the run.

### 5.1 Experiment 1: Position trajectory tracking

The arm tracks a closed end-effector position trajectory; the tip attitude (angle $\psi$) is not commanded. Both controllers follow the position reference closely, but the proposed method keeps the uncommanded attitude bounded, whereas the PCC baseline exhibits a large $\psi$ excursion near the mid-trajectory configuration. The proposed position errors remain near zero across the cycle; the baseline shows transient error at the trajectory extremes. The online estimated Jacobian error stays essentially flat for the proposed method but accumulates and oscillates for the PCC baseline, indicating progressive model mismatch that the compensation removes (**Figure 3**).

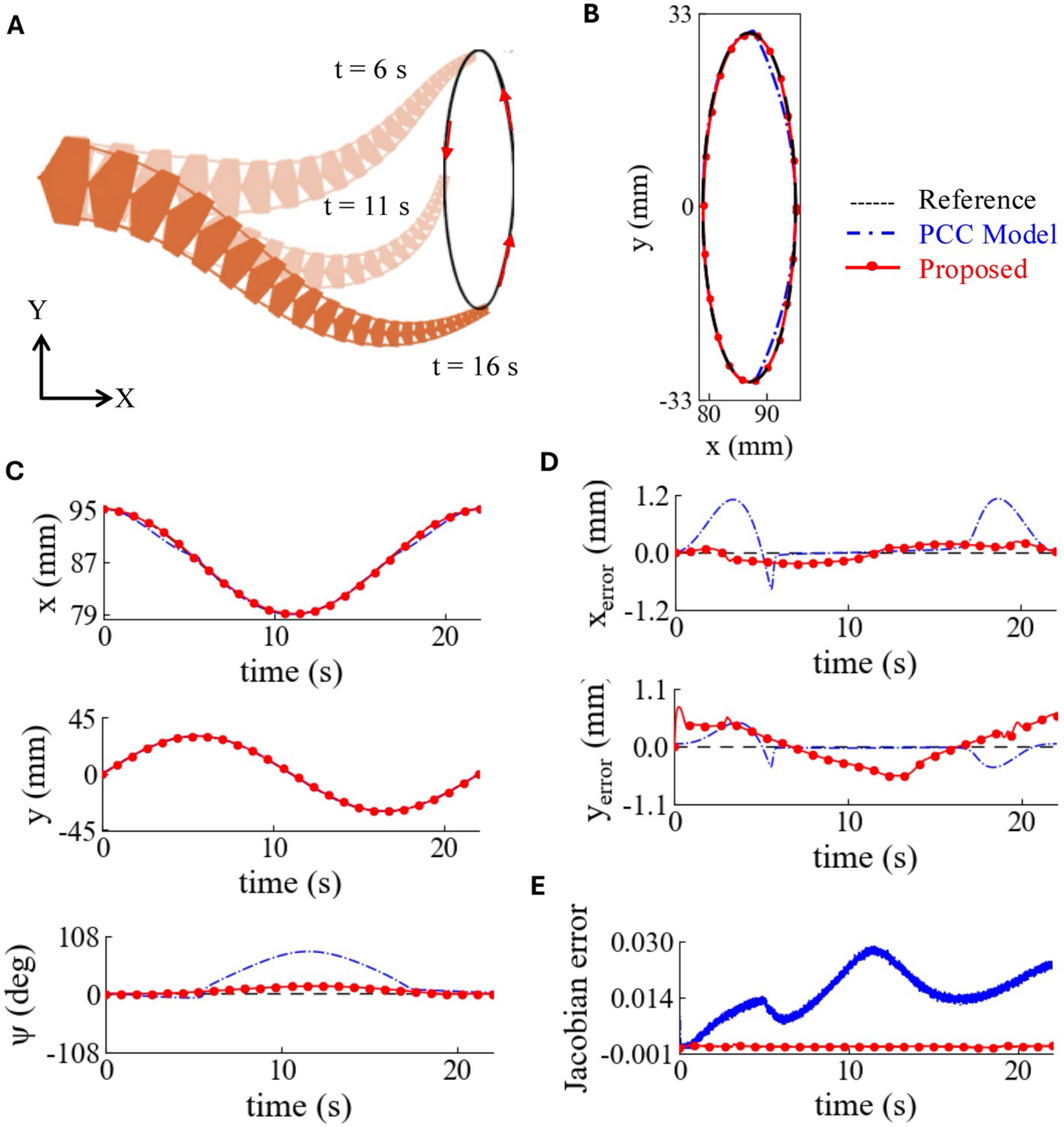

**Figure 3. Trajectory tracking at fixed attitude. (A)** Representative configurations of the logarithmic-spiral continuum arm during closed-loop tracking. **(B)** End-effector trajectory in task space, comparing the reference path, PCC baseline, and proposed controller. **(C)** Time histories of end-effector position $(x, y)$and attitude $\psi$. **(D)** Position-tracking errors in x and y. **(E)** Online Jacobian-error metric. The proposed morphology-specific controller maintains accurate trajectory tracking, small attitude deviation, and near-zero Jacobian error, whereas the PCC baseline exhibits increased attitude drift and accumulated modeling error.

**Table 2.** End-effector tracking errors for Experiment 1 (RMSE / MAE).

| Method | x (mm) | y (mm) | ψ (deg) |
|---|---|---|---|
| **Proposed** | 0.1486 / 0.1329 | 0.3445 / 0.2997 | 8.0272 / 5.9962 |
| PCC baseline | 0.4744 / 0.2998 | 0.1787 / 0.1208 | 41.4919 / 30.1973 |
| Open-loop | 13.5329 / 10.7262 | 27.3271 / 22.9260 | 2.6723 / 2.1808 |

### 5.2 Experiment 2: Attitude regulation with stationary position

The attitude $\psi$ is commanded over a range (approximately $-30°$ to $+30°$) while the position is held stationary, thereby isolating convergence from a large initial error. The proposed method drives the error to near zero within roughly one to two seconds, while the PCC baseline converges over the full horizon with a slowly decaying transverse oscillation. The estimated Jacobian error remains flat for the proposed method and grows for the baseline, explaining the sluggish, oscillatory regulation of the uncompensated controller (**Figure 4**).

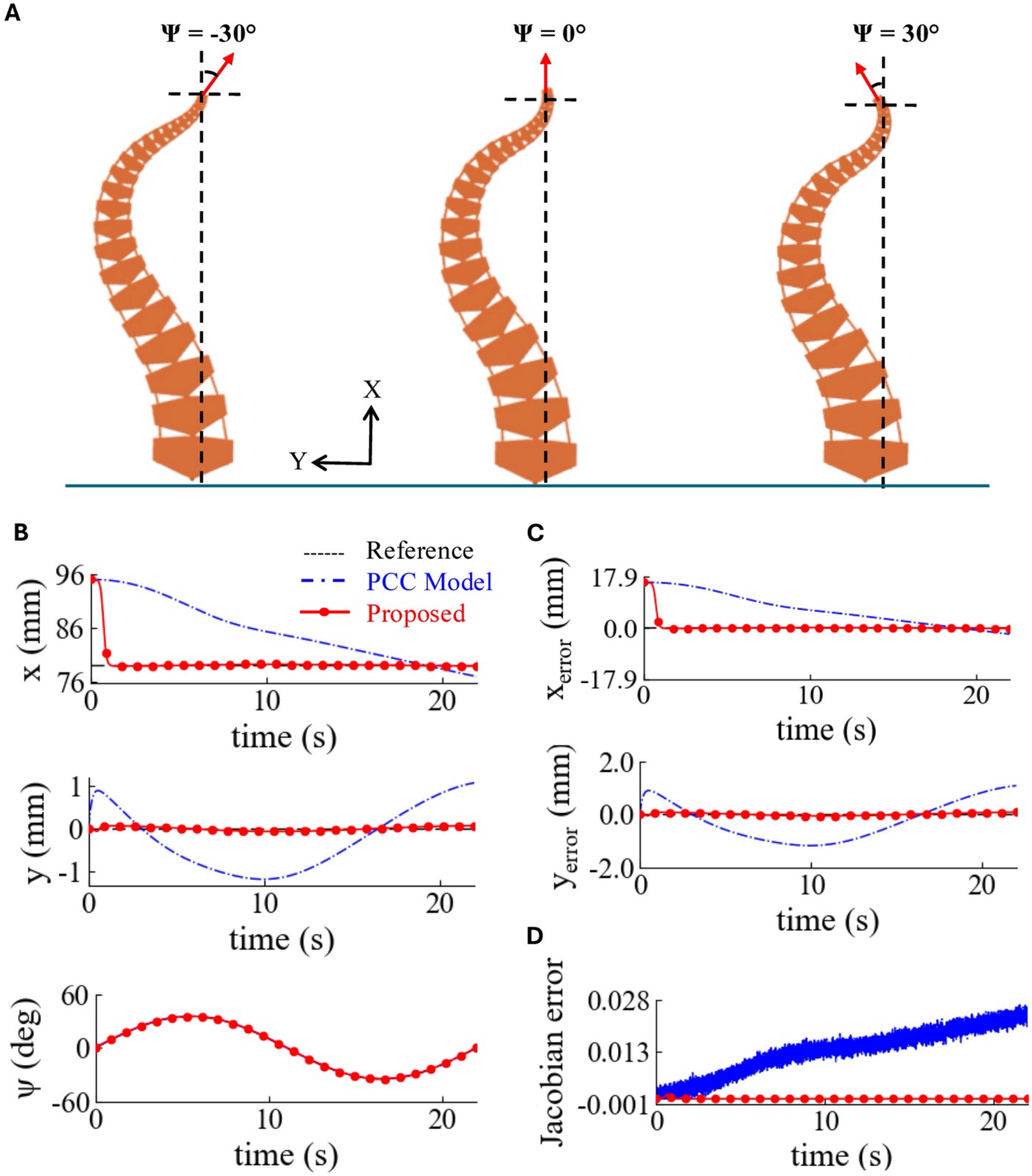

**Figure 4. Attitude regulation with stationary position. (A)** Representative configurations of the logarithmic-spiral continuum arm during attitude regulation at $\psi = -30°, 0°$,and $+30°$, while maintaining a fixed end-effector position. **(B)** Time histories of end-effector position $(x, y)$and attitude $\psi$ for the reference trajectory, PCC baseline, and proposed controller. **(C)** Position-tracking errors in x and y. **(D)** Online Jacobian-error metric. The proposed morphology-specific controller rapidly converges to the desired attitude while maintaining the target position and a near-zero Jacobian error, whereas the PCC baseline exhibits slower convergence, larger position deviation, and accumulated Jacobian error.

**Table 3.** End-effector errors for Experiment 2 (RMSE/MAE) and convergence time.

| Method | x (mm) | y (mm) | ψ (deg) | $t_{conv}$ (s) |
|---|---|---|---|---|
| **Proposed** | 2.7682/0.6236 | 0.0474/0.0426 | 0.0046/0.0036 | 0.9505 |
| PCC baseline | 8.5268/6.7605 | 0.7779/0.6933 | 0.0353/0.0311 | 11.6803 |

### 5.3 Experiment 3: Tracking under external disturbances

External disturbances are introduced while the arm tracks the reference under the proposed controller. Disturbances appear as transient spikes in the position errors at the instants of application; the controller rejects each and returns the tip to the path, with attitude error remaining bounded. The estimated Jacobian error spikes briefly at each disturbance and settles back toward zero, showing that the online compensation absorbs the unmodeled interaction and restores accurate tracking (**Figure 5**). This robustness is essential for grasping, where contact and payload vary continuously.

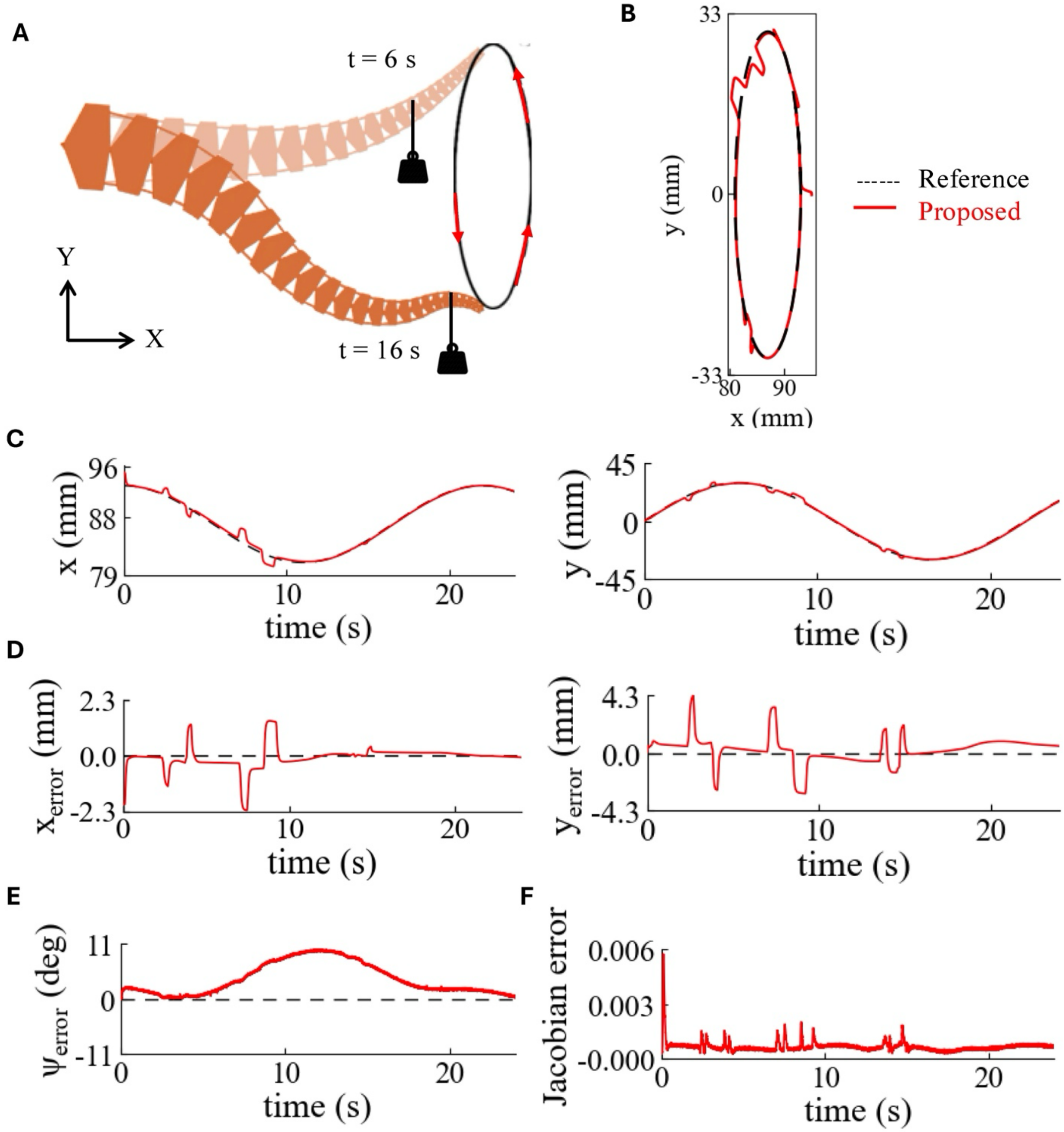

**Figure 5. Trajectory tracking under external disturbances. (A)** Representative configurations of the logarithmic-spiral continuum arm during closed-loop trajectory tracking with external disturbances applied at t=6 s and t=16 s. **(B)** End-effector trajectory compared with the reference path. **(C)** Time histories of end-effector position $(x, y)$. **(D)** Position-tracking errors in $x$ and $y$, showing transient deviations caused by the disturbances and subsequent recovery. **(E)** Attitude-tracking error $\psi_{\text{error}}$. **(F)** Online Jacobian-error metric. The proposed controller rapidly compensates for disturbance-induced deviations, maintains bounded attitude error, and preserves a near-zero Jacobian error throughout the task.

**Table 4.** Disturbance response (maximum absolute error, recovery time, post-disturbance RMSE).

| Disturbance | post-RMSE / MaxAE **x** (mm) | post-RMSE / MaxAE **y** (mm) | MaxAE **ψ** (deg) | Average Recovery / |
|---|---|---|---|---|
| Adding weight | 0.1975 / 0.1545 | 0.4992/ 0.4273 | 5.1004/ 3.9690 | 0.6980 s |

**5.4 Experiment 4: 3D tracking with the three-tendon arm**

Experiment 4 evaluates a spatial arm (three equilateral cables per segment) that reaches out-of-plane targets and has a four-degree-of-freedom task budget ($x, y, z, \theta$; Section 2.1). Within this budget, we pose two well-conditioned three-output subtasks, each leaving one degree of freedom redundant. Case 4A tracks the full 3D tip position ($x, y, z$), leaving tip orientation free. Case 4B tracks two position coordinates ($\mathrm{x}, \mathrm{y}$) together with the tip elevation angle θ (the inclination of the backbone tangent at the tip above the global xy-plane), leaving the vertical coordinate z and the tip azimuth free. Each subtask uses a wide 3×4 task Jacobian, and redundancy is resolved by the damped least-squares update of Eq. (18). Together, the cases show that the arm can allocate three of its four controllable degrees of freedom either entirely to position or to a mix of position and pointing. The corresponding trajectories and errors are reported in Tables 5A–5B.

In Case 4A, the proposed controller is expected to track the spatial position reference with small, bounded error and a flat estimated Jacobian-error trace, whereas the PCC baseline accumulates Jacobian error and position drift. In Case 4B, the controlled outputs are $x$, $y$, and $\theta$. Because z and azimuth are deliberately left free, the arm may drift along these uncontrolled directions, and such drift is expected rather than a tracking failure. Controlling elevation rather than azimuth avoids the heading degeneracy that arises when the tip tangent approaches vertical.

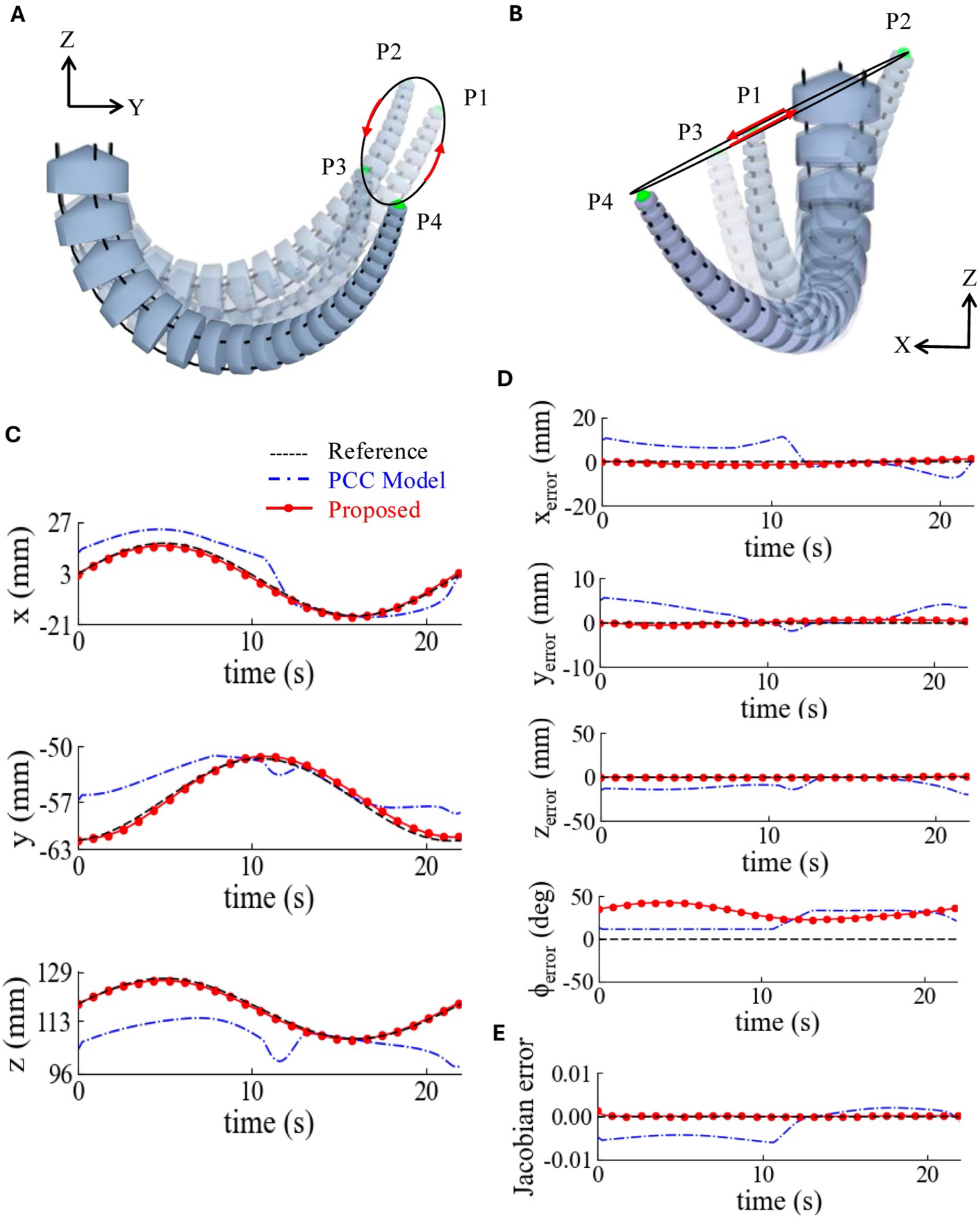

**Figure 6. Three-dimensional task-space tracking of the spatial logarithmic-spiral continuum arm. (A)** Representative arm configurations during spatial position tracking, showing the evolution of the end-effector trajectory through four reference points (P1–P4). **(B)** Side view of the corresponding out-of-plane motion. **(C)**

Time histories of the end-effector position $(x, y, z)$ for the reference trajectory, PCC baseline, and proposed controller. **(D)** Tracking errors in $x$, $y$, and $z$, together with the tip-elevation error $\phi$. **(E)** Online Jacobian-error metric. The proposed morphology-specific controller accurately tracks the three-dimensional reference trajectory while maintaining a near-zero Jacobian error, whereas the PCC baseline exhibits substantial tracking deviations and accumulated modeling error due to geometric mismatch.

**Table 5A.** Full 3D position tracking errors (RMSE/MAE); tip orientation uncontrolled.

| Method | x (mm) | y (mm) | z (mm) |
|---|---|---|---|
| **Proposed** | 0.9907/ 0.8618 | 0.4785 / 0.4284 | 0.5628 / 0.4957 |
| PCC baseline | 6.2468/5.2908 | 2.9337/2.3142 | 10.6087/9.1299 |
| Open-loop | 11.3243 / 10.1615 | 21.5857 / 17.2214 | 45.5944 / 33.4223 |

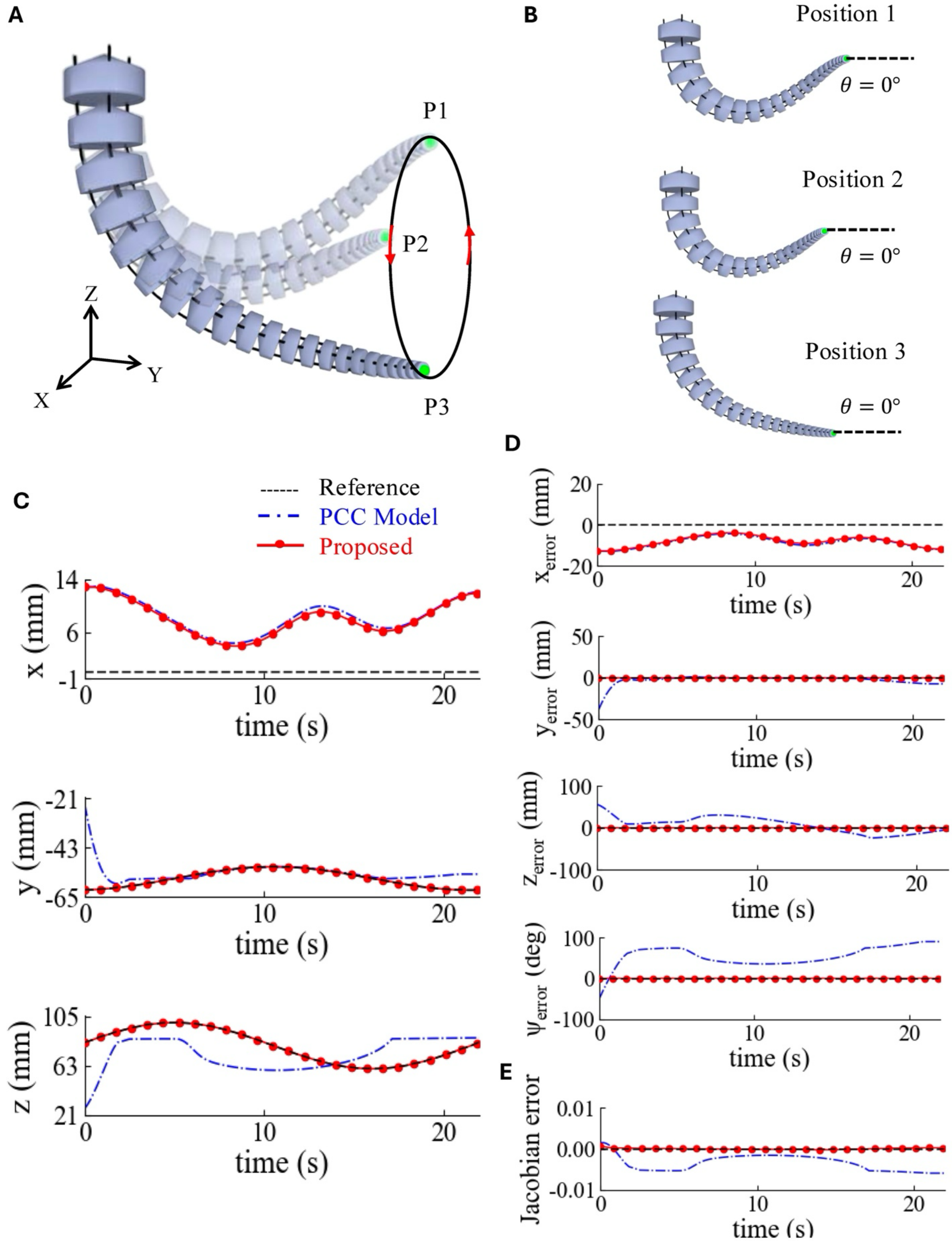
A
B
P1
P2
P3
Z
Y
X
Position 1
Position 2
Position 3
$\theta = 0°$
$\theta = 0°$
$\theta = 0°$
C
Reference
PCC Model
Proposed
x (mm)
y (mm)
z (mm)
time (s)
D
$x_{error}$ (mm)
$y_{error}$ (mm)
$z_{error}$ (mm)
$\psi_{error}$ (deg)
E
Jacobian error

**Figure 7. Spatial trajectory tracking with end-effector orientation regulation. (A)** Representative configurations of the logarithmic-spiral continuum arm during three-dimensional trajectory tracking through reference points P1–P3. **(B)** Snapshots of the arm at different locations along the trajectory, demonstrating maintenance of a prescribed end-effector elevation angle ($\theta = 0°$) throughout the motion. **(C)** Time histories of end-effector position (*x,y,z*) for the reference trajectory, PCC baseline, and proposed controller. **(D)** Tracking errors in $x, y,$ and $z$, together with the elevation-angle error $\theta_{\mathbf{error}}$. **(E)** Online Jacobian-error metric.

**Table 5B.** Position ($\mathrm{x}, \mathrm{y}$) and tip elevation $\theta$ tracking errors (RMSE/MAE); $\mathrm{z}$ and azimuth uncontrolled.

| Method | x (mm) | y (mm) | θ (deg) |
|---|---|---|---|
| **Proposed** | 8.3991/8.0021 | 0.0022 / 0.0019 | 0.0114/0.0104 |
| PCC baseline | 8.7695/8.4085 | 5.6042/2.7011 | 60.0486/56.5868 |
| Open-loop | 20.2811 / 16.2250 | 38.1298/ 32.9664 | 47.3969 / 40.7555 |

### 5.5 Grasping demonstrations: closed-loop reaching with morphology-driven wrapping

These demonstrations exercise the division of labor introduced in Section 4: the closed-loop controller positions and orients the arm so that the target falls within the spiral's enveloping range, after which wrapping and holding are produced by the arm's passive morphology rather than by feedback. We do not regulate contact forces or contact locations—quantities that are unobservable on this platform and far exceed the arms at most four controllable task degrees of freedom—and instead report grasp outcomes as evidence of morphological (mechanical) intelligence. We demonstrate grasping in three settings: a two-tendon arm that reaches a target while curling around an obstacle; a single three-tendon arm performing enlarged reach, approach, and whole-arm wrapping to envelop a compact object; and a coordinated multi-arm gripper performing a full pick-and-place cycle on a large object (rest, approach, grasp, pick-up, replace). The same task-space framework scales from single-arm to multi-arm entanglement grasps using only approximate object location (Figure 8). Across the three settings, we report the number of trials, success rate, object size and mass range, and the dominant failure modes (**Table 6**); a grasp is scored as successful when the arm reaches, encloses, and holds the object through the lift without slip. These demonstrations are intended as evidence of feasibility for morphological intelligence rather than as precision-control results: the equiangular spiral curls into a geometrically self-similar, scale-invariant envelope, so a single coarse reaching command can adapt to objects of different sizes without per-configuration retuning or contact-level feedback. The controller's role is confined to placing the object within this enveloping range, where reaching is the limiting factor, grasp success follows the tracking accuracy quantified in Experiments 1 and 4, rather than any regulated contact wrench.

For the grasping experiment in MuJoCo, we selected 3 positions within the arm workspace. For each position, we conducted 3 random trials, totaling 9 trials for each arm type. The object size and mass differ; the measurements are recorded in **Table 6**. In the table, D means diameter and H means height for the object size. For the two-tendon and three-tendon single arms, we selected a cylindrical shape for stability, and a spherical shape was used for the multi-array arm. In the simulation, we randomized the object position for grasping and recorded the data. The two-tendon single arm successfully grasped 6 out of 9 trials, whereas the three-tendon single arm successfully grasped 8 out of 9 trials. The dominant failure mode of the two-tendon arm is an out-of-plane reach issue. Since this is a planar version, the arm cannot reach all locations that are out-of-plane. Subsequently, the three-tendon arm can reach more positions than the two-tendon arm. Therefore, its reach is better than the other one, but there are still issues with limited contact coverage. So, this version cannot grasp larger or differently shaped objects. To report on the performance of the multi-array arm, 7 out of 9 trials were successful. The two unsuccessful cases were due to the position of the arm base center. The dominant failure mode for this version remains the correct center-aligned position with the object; otherwise, the load is unstable for pick-and-place applications.

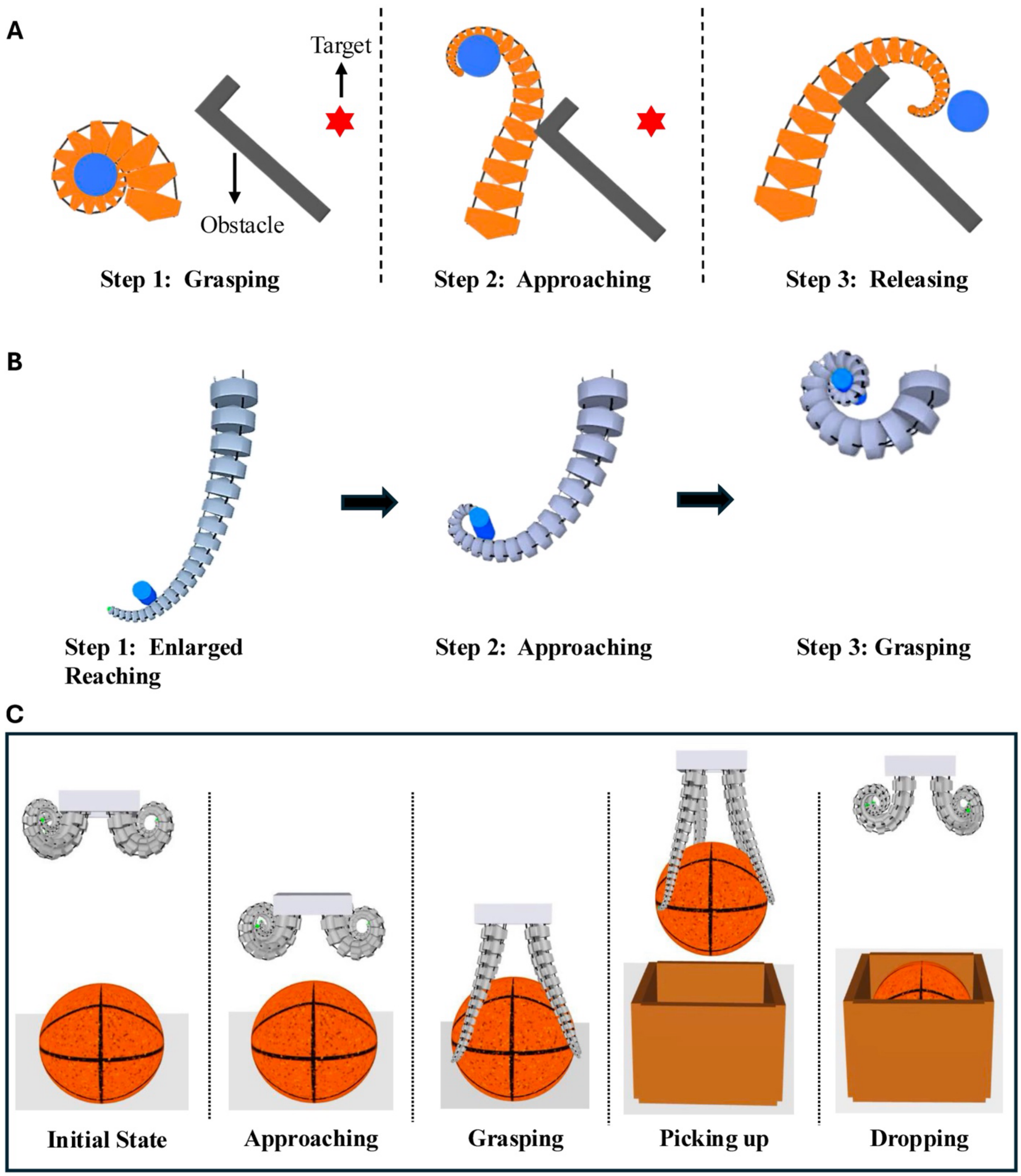

**Figure 8. Morphology-enabled manipulation behaviors of logarithmic-spiral continuum arms. (A)** Reach–wrap–release manipulation around an obstacle. The arm first grasps an object, transports it toward a target location while avoiding the obstacle, and subsequently releases the object at the desired destination. **(B)** Morphology-enabled enlarged reaching and adaptive grasping. The arm initially extends into an enlarged configuration to increase reachability, then curls to approach and securely grasp the target object. **(C)**

Cooperative manipulation using a dual-arm logarithmic-spiral system. Two continuum arms coordinate to approach, grasp, lift, transport, and place a spherical object into a container, demonstrating the potential for bimanual manipulation and object handling.

**Table 6.** Grasping performance across the three demonstration settings.

| Grasp configuration | Object size (mm) | Object mass (g) | Trials | Success rate (%) | Dominant failure mode |
|---|---|---|---|---|---|
| Two-tendon, reach around obstacle | 13 (D)<br>13 (H) | 220 | 9 | 6/9 | Out-of-plane reach, 6 out of 9 trials were successful |
| Single three-tendon, whole-arm wrap | 15 (D)<br>13 (H) | 300 | 9 | 8/9 | Object size-limited enclosure, contact coverage is insufficient |
| Multi-arm array, pick-and-place | 70 (D) | 1300 | 9 | 7/9 | Highly dependent on the arm closure center, uneven center location results in uneven load to pick |

## 6. Discussion

### 6.1 Biological Implications of Logarithmic-Spiral Morphology

Logarithmic spirals are common in biological systems, including octopus arms, elephant trunks, plant tendrils, chameleon tails, and climbing vines, where they enable efficient reaching, wrapping, anchoring, and object manipulation. Unlike structures with constant curvature, these natural appendages often show continuously varying curvature and diameter along their length, allowing them to adapt their morphology to diverse environmental constraints and interaction tasks. This geometric variation enables a unique combination of compact storage, a large reachable workspace, and efficient whole-body grasping.

The present study suggests that the logarithmic-spiral geometry offers both mechanical and control advantages. Because curvature varies smoothly along the arm, the resulting shape naturally aligns with many reach–wrap–hold manipulation tasks observed in biological systems. The ability to transition continuously between extended and tightly curled configurations allows the manipulator to achieve large changes in workspace coverage without additional joints or actuators. These observations support the hypothesis that spiral morphologies may represent an effective evolutionary solution for balancing dexterity, compliance, and control efficiency in highly deformable appendages.

### 6.2 Morphological Intelligence and Control Simplification

An emerging concept in soft robotics is *morphological intelligence*, in which mechanical design and geometry contribute directly to task execution, thereby reducing the burden on sensing, planning, and control. Rather than treating morphology as a passive structure that must be compensated for by increasingly sophisticated controllers, morphology can actively shape the control landscape and simplify the underlying control problem.

The logarithmic-spiral arm investigated here is a clear example of morphology-assisted control. Because the arm naturally curls into a spiral configuration, many manipulation behaviors, such as enclosing an object, wrapping around obstacles, or transitioning between reaching and grasping states, arise from its intrinsic geometry rather than requiring complex trajectory generation. The controller therefore regulates motion within a morphology-constrained configuration space that is already well suited to the target tasks.

This concept is illustrated by the manipulation examples in Fig. 8, where reaching, wrapping, obstacle avoidance, and whole-arm grasping emerge naturally from the arm's geometry. In contrast, achieving comparable behaviors with conventional continuum manipulators often requires more complex shape planning and higher-dimensional control strategies. Consequently, the proposed framework shows that morphology-specific modeling can leverage geometric intelligence to reduce control complexity while preserving task performance.

### 6.3 Comparison with PCC and Cosserat-Based Approaches

Most continuum robot control methods rely on either PCC models or Cosserat-rod formulations. PCC models remain attractive for their simplicity and computational efficiency; however, they assume constant curvature within each section and therefore cannot accurately represent geometries with continuously varying curvature along the backbone. For logarithmic-spiral manipulators, this mismatch introduces geometric approximation errors that propagate into the Jacobian, ultimately degrading closed-loop tracking performance.

The results in Figs. 3–7 show that these errors become increasingly pronounced during large deformations, spatial motions, and orientation regulation tasks. In contrast, the proposed logarithmic-spiral formulation derives the kinematics and Jacobians directly from the underlying morphology, thereby preserving the arm’s continuously varying curvature. As a result, the controller exhibits reduced tracking error, less attitude drift, and improved robustness to disturbances compared with the PCC baseline.

At the opposite end of the modeling spectrum, Cosserat-rod formulations provide highly accurate representations of continuum robots and can capture distributed deformation, material nonlinearities, and external loading. However, these models typically require numerical integration and often incur substantially higher computational costs than analytical approaches. Real-time control based directly on Cosserat models can therefore be challenging, particularly for highly compliant systems with significant contact interactions.

The proposed framework sits between these two extremes. It retains much of the computational efficiency and analytical tractability of reduced-order models while incorporating morphology-specific geometric information absent from conventional PCC formulations. Consequently, the approach offers a practical compromise between model fidelity and real-time control capability for logarithmic-spiral manipulators.

### 6.4 Limitations and Future Directions

Several limitations of the present study should be acknowledged. First, all validation was performed in simulation using the MuJoCo physics engine. Although the simulator captures tendon actuation, contact interactions, and nonlinear deformation, discrepancies between simulation and physical hardware are inevitable. Future work will focus on building a physical platform and experimentally validating the proposed controller under realistic sensing uncertainties, actuator nonlinearities, tendon friction, hysteresis, and external disturbances.

Second, the current framework is primarily kinematic and quasi-static. Dynamic effects, including inertia, damping, viscoelasticity, tendon compliance, and impact interactions, are not explicitly modeled. Extending the framework to incorporate dynamic modeling and control will be essential for high-speed manipulation and locomotion tasks.

Third, the online Jacobian estimator relies on local updates computed from measured motion and actuation data. While the proposed Broyden–Kalman approach effectively compensates for moderate modeling errors, performance may degrade under severe unmodeled dynamics, abrupt topology changes, or large contact-induced discontinuities. Future work may integrate physics-informed machine learning and adaptive estimation techniques to improve robustness in these conditions.

Finally, the current study considered a limited set of benchmark tasks and manipulation scenarios. Future investigations will explore cooperative multi-arm manipulation, contact-rich interactions, autonomous grasp planning, obstacle-rich environments, and learning-augmented control. In particular, combining morphology-specific analytical models with data-driven residual learning may offer a promising path toward scalable, real-time control of highly deformable continuum manipulators in complex environments.

## 7. Conclusion

This paper presents the first morphology-specific closed-loop task-space control framework for logarithmic-spiral continuum manipulators. An analytical Jacobian derived directly from logarithmic-spiral kinematics is combined with online Broyden-based adaptation and Kalman-filter Jacobian error compensation to address modeling uncertainty, contact effects, and geometric mismatch. The resulting controller enables accurate and robust control of highly underactuated spiral continuum arms while preserving the geometric characteristics of the underlying morphology.

Comprehensive simulation studies, including trajectory tracking, attitude regulation, disturbance rejection, three-dimensional position and position-orientation tracking, and morphology-enabled manipulation tasks, demonstrate consistent improvements over PCC and open-loop baselines. The framework also enables closed-loop reaching behaviors that transition naturally into reach–wrap–hold grasping, including obstacle-assisted manipulation, whole-arm wrapping, and cooperative multi-arm entanglement grasping. These results address the feedback-control limitation identified in prior SpiRobs studies and establish a control foundation for logarithmic-spiral robotic systems.

More broadly, this work highlights the importance of morphology-aware modeling and control in continuum robotics, demonstrating how geometric intelligence embedded in the spiral architecture can simplify manipulation and improve performance. Future work will focus on experimental validation on physical SpiRobs platforms, dynamic modeling and control, contact-aware manipulation, and learning-augmented hybrid controllers for operation in complex and unstructured environments.

**Supplementary Material.** Model-validation results (Supplementary Section S1 and Supplementary Figure S1) and detailed statics, compliance, and task-Jacobian conditioning derivations (Supplementary Section S2) are provided in the online Supplementary Material.

### Acknowledgements

This work is supported by the National Science Foundation (ECCS-2024649) and Case Western Reserve University.

## Supplementary Material for

## Morphology-Specific Closed-Loop Control of Logarithmic-Spiral Continuum Arms via Online Jacobian Error Compensation

Partha Datta[1,2,†], Yi Jin[1,†], Wei Lin[2], C. Chase Cao[1,2,3,*]

[1]*Laboratory for Soft Machines and Electronics, Department of Mechanical and Aerospace Engineering, Case Western Reserve University, Cleveland, OH 44106, USA*

[2]*Department of Electrical, Computer and Systems Engineering, Case Western Reserve University, Cleveland, OH 44106, USA*

[3]*Advanced Platform Technology (APT) Center, Louis Stokes Cleveland VA Medical Center, Cleveland, OH 44106, USA*

†Equal contribution. *Corresponding author: C. Chase Cao (ccao@case.edu)

### S1. Model validation: bending and stiffness profile

To confirm that the segmented model preserves continuum mechanics, the bending profile of the MuJoCo arm is compared with an analytical Cosserat-rod model of the same logarithmic spiral, and per-segment stiffness is examined. The simulated backbone reproduces the analytical spiral, and both joint and equivalent bending stiffnesses decrease monotonically from base to tip, consistent with the graded compliance imposed by the taper (Fig. S1).

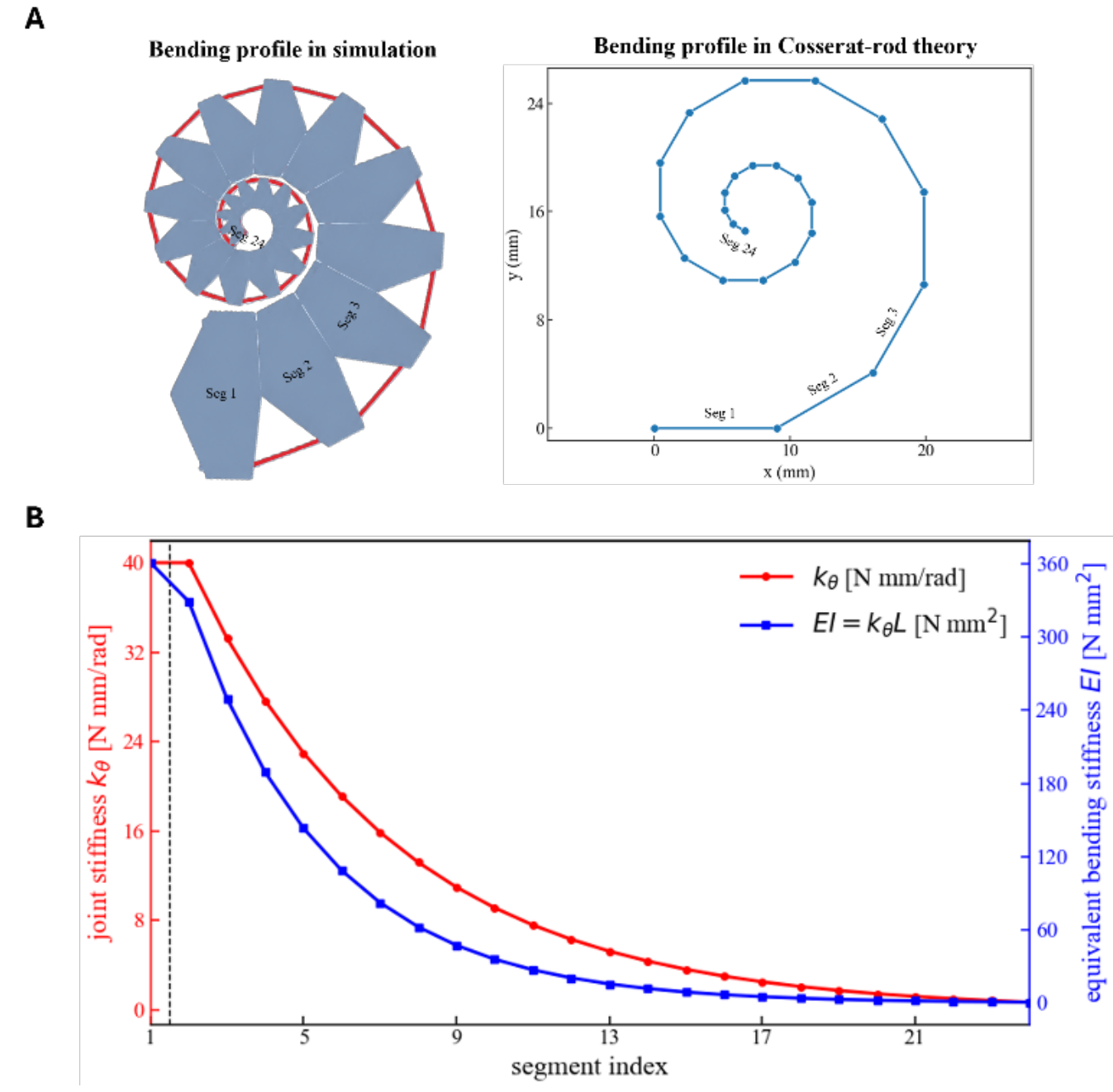

**Supplementary Figure S1.** Compliance characterization. (A) Bending profile in simulation and in the analytical Cosserat-rod model. (B) Effective bending stiffness EI across the segment index. (C) Joint stiffness $k\theta$ and equivalent bending stiffness for each segment.

**S2. Detailed Statics, Compliance, and Task-Jacobian Conditioning**

This appendix presents the static and differential relations underlying the compliance and feasibility claims in Sections 2 and 5. The quasi-static kinematic controller in Section 3 does not require them, but they characterize the load response and the conditioning of the task Jacobian. Let $\Theta = \{\theta_{i,j}\} \in \mathbb{R}^{2m}$ ($M = \Sigma_i N_i$) collect all joint bending vectors, with block stiffness $K = \text{diag}(k_{i,j} I_2)$ and elastic energy $U = ½ \Sigma k_{i,j} \|\theta_{i,j}\|^2$. The cable generalized force is $\tau_{i,j} = d_{i,j} M_i$, and an external tip wrench W enters via the geometric Jacobian $J_g = \partial(\text{tip})/\partial\Theta \in \mathbb{R}^{6\text{x}2m}$. At equilibrium, the tangent stiffness and the tip compliance are

$$-K\Theta + \tau + J_g^\top W = 0, \qquad K_T = K - K_G, \qquad C = J_g K_T^{-1} J_g^\top$$

where the geometric stiffness is $K_G = \partial(J_g^\top W)/\partial\Theta$ and saturated joints ($\|\theta_{i,j}\| = \beta_i$) are treated as rigid, so they drop out of $K_T^{-1}$. At $W = 0$, this recovers the unloaded profile $\theta_{i,j} = (d_{i,j}/k_{i,j}) M_i$ of Eq. (3). Because $k_{i,j} \propto \gamma_i^{3j}$ is smallest at the distal joints, the tip compliance C is tip-dominated; and because saturated joints leave $K_T^{-1}$, the structural compliance decreases as the arm curls (self-stiffening through saturation). The load-coupled task Jacobian is $J_q^{\text{loaded}} = D K_T^{-1} K B$, with $D = \partial x/\partial\Theta$ and $B = \partial\Theta_{\text{cmd}}/\partial q$.

The per-segment tendon Jacobian relating curvature-vector changes to cable-length changes is

$$J_{t,i} = \frac{\partial \ell_i}{\partial u_i} = -C_i \begin{bmatrix} \cos\psi_{i,1} & \sin\psi_{i,1} \\ \cos\psi_{i,2} & \sin\psi_{i,2} \\ \cos\psi_{i,3} & \sin\psi_{i,3} \end{bmatrix}, \quad C_i = d_{i,0} \sum_j \gamma_i^{-j}$$

which is constant in the curvature-vector coordinates and of rank 2, so all conditioning variation resides in the task Jacobian $J_q$ rather than in the tendon map. Feasibility is read from the singular values $\sigma_1 \geq \cdots \geq \sigma_d$ of the length-normalized $J_q$ (the elevation row scaled by a characteristic length), via the manipulability $w = \prod_k \sigma_k$ and the condition number $\varkappa = \sigma_1/\sigma_d$. For the two-segment arm, the rank drops on the coplanar manifold $\varphi_1 \equiv \varphi_2$ (mod $\pi$) and as $A_i \to \beta_i$ (saturation); these are the two boundaries that define well-posed four-degree-of-freedom control, which is why the 3D tracking cases of Section 5 use three-output subtasks that retain a redundant degree of freedom.